\documentclass[journal,10pt]{IEEEtran}

\usepackage{multicol}
\usepackage{graphicx}
\usepackage{amsmath}
\usepackage{cuted}
\usepackage{cite}
\usepackage{multirow}
\usepackage{hhline}
\usepackage{xcolor}
\usepackage{balance}
\usepackage{subfigure}
\usepackage[switch]{lineno}
\usepackage{stfloats}

\usepackage{algorithm}
\usepackage{algpseudocode}
\usepackage{soul}
\usepackage[normalem]{ulem}

\definecolor{newblue}{RGB}{8,2,162}

\makeatletter
\def\hlinewd#1{%
\noalign{\ifnum0=`}\fi\hrule \@height #1 \futurelet
\reserved@a\@xhline}

\makeatother
\begin{document}

\title{A Pedestrian-Sensitive Training Algorithm for False Positives Suppression in Two-stage CNN Detection Methods}

\author{Qiang Guo

\thanks{Qiang Guo are with College of Mechanical and Electronic Engineering, Dalian Minzu University, Dalian 116024, China
(e-mail: guoqiang0148666@gmail.com).}
}

\markboth{IEEE TRANSACTIONS ON INSTRUMENTATION AND MEASUREMENT,~Vol.~XX, No.~XX, XXX~2023}
{}

\maketitle

\begin{abstract}
Pedestrian detection has been a hot spot in computer vision over the past decades due to the wide spectrum of promising applications, the major challenge of which is False Positives (FPs) that occur during pedestrian detection. The emergence various Convolutional Neural Network-based detection strategies substantially enhance the pedestrian detection accuracy but still not well solve this problem. This paper deeply analysis the detection framework of the two-stage CNN detection methods and find out false positives in detection results is due to its training strategy miss classify some false proposals, thus weakens the classification capability of following subnetwork and hardly to suppress false ones. To solve this problem, This paper proposes a pedestrian-sensitive training algorithm to effectively help two-stage CNN detection methods learn to distinguish the pedestrian and non-pedestrian samples and suppress the false positives in final detection results. The core of the proposed training algorithm is to redesign the training proposal generating pipeline of the two-stage CNN detection methods, which can avoid a certain number of false ones that mislead its training process. With the help of the proposed algorithm, the detection accuracy of the MetroNext, an smaller and accurate metro passenger detector, is further improved, which further decreases false ones in its metro passengers detection results. Based on various challenging benchmark datasets, experiment results have demonstrated that feasibility of the proposed algorithm is effective to improve pedestrian detection accuracy by removing the false positives. Compared with the existing state-of-the-art detection networks, MetroNext-PST demonstrates better overall prediction performance in accuracy, total number of parameters, and inference time, thus it can become a practical solution for hunting pedestrian on various hardware platforms, especially tailored for mobile and edge devices.
\end{abstract}


\IEEEpeerreviewmaketitle


\section{Introduction}

\IEEEPARstart{P}{edestrian} detection has always been a fundamental technique of various artificial intelligence tasks, such as autonomous driving, pedestrian tracking, abnormal behavior detection, etc., and played an indispensable role in their successful applications and deployments \cite{Ref1}. However, compared to the general object detection, pedestrian detection has its specific technical difficulties that make it the most challenging subfield of object detection.

First, pedestrians are in complex and ever-changing scenes, wherein some other objects have similar appearances to the pedestrians, which lead to the detector can't differentiate such objects even by pedestrian feature extraction. Secondly, the high intra-class variation of pedestrian in clothing, lighting and pose, etc., requires the learned human features to be more semantically meaningful and robust to achieve accurate pedestrian recognition, which brings a big challenge for the design of feature extraction and detection strategies. Finally, The pedestrians are always occluded by other objects or each others, and only partial human bodies can be observed by the detectors, thus failing to provide the detector with enough pedestrian features to segment and localise individual pedestrian from the crowd. Therefore, over the past decades, a major methodology is to enrich and acquire key pedestrian features for recognising pedestrians in complex scenes. \cite{Ref2,Ref3}. Recently, deep learning and in particular Convolutional Neural Network (CNN) has emerged as the preferred algorithm for pedestrian detection. Aided by advances in the realm of CNN, significant improvements have been witnessed in this field. \cite{Ref3}. However, the False Positives (FPs), a notorious problem in pedestrian detection, is still not well solved because although the use of CNN has enhanced the feature extraction capability of the whole detection framework, the most CNN-based pedestrian detectors are transferred from the CNN-based general object detectors, which emphasis target position awareness and multiple object classification, but the pedestrian detection belongs to a binary classification problem, which requires the established detector has pedestrian-awareness ability and can be effectively classify pedestrian and non-pedestrian objects and avoid FPs from being incorporated into the detection results, which is negligible problem in the field of pedestrian detection and causes the CNN-based detectors are not the optimal solutions for detecting pedestrians and their pedestrian detection performance are also difficult to further improve.


%
%

%
In order to solve the above problems and achieve better detection accuracy, this paper deeply analysis the detection pipeline of the accurate-oriented two-stage CNN-based detection network and points out that its training strategies can't help the whole detection network has stronger classification ability to distinguish pedestrian and non-pedestrian to achieve wipe out FPs. To tackle this problem, a novel Pedestrian-Sensitive Training (PST) algorithm is proposed to enhance the classification ability of the two-stage CNN-based detection network and help it overcome this deficiency and eliminate these FPs. In addition, the demand of recognizing pedestrian objects are mainly taken placed in edges devices, which inevitable requires the proposed solution can improve the FPs removing ability of the plain detector at lower or without extra computational cost, satisfying the hardware constraints of the targets platforms. In view of these limitations, this paper chooses a small two-stage detector MetroNext \cite{Ref4}, and then combines it with the PST algorithm to establish a small but fast and accurate pedestrian detector MetroNext-PST. Finally, the MetroNext-PST is validated on a real-life metro station datasets: SY-Metro and an embedded platform to test whether it can be used to fast and accurate detect metro passengers to replace human surveillance. To sum up, the contributions of this paper are three-fold.

\begin{itemize}
  \item [1)]
  this paper deeply analysis the detection pipeline of the accurate-oriented two-stage CNN-based detection network and points out that its training strategies can't help the whole detection network has stronger classification ability to distinguish pedestrian and non-pedestrian. Thus, a novel Pedestrian-Sensitive Training algorithm is proposed, which can effectively guide the training process of the two-stage CNN-based detection network and promotes its classification capability to wipe out non-pedestrian predictions. The PST algorithm only works on the training stage of the two-stage CNN-based detection method and not brings extra computational cost in its inference stage, thus achieving zero-cost increases in the prediction accuracy of the plain detection network.
  \item [2)]
  Various experiments have been conducted on the challenging benchmark datasets including CUHK Occlusion (CUHK-Occ), Caltech and CityPersons and a real-life metro station datasets to fully validate the feasibility of the proposed methods. Firstly, the ablation experiments on CityPersons dataset demonstrate the general capability of the PST algorithm to enhance the prediction accuracy of the plain model without extra computation cost at the inference stage. Then, combining the PST algorithm with the small detection network: MetroNext, a newly-built MetroNext-PST is tesed on the above benchmark datasets, which shows that the PST algorithm has good adaptability and can effectively enhance the classification ability of pedestrian recognition even for small detection models, effectively removing erroneous positive examples. At the same time, with the help of PST algorithm, it obtains good comprehensive pedestrian detection performance under small model size, which shows that it is a practical model for solving the pedestrian detection problem in embedded scenes. Finally, the metro scene dataset is used to simulate the test of PTSNet's ability to accurately identify the pedestrian targets in crowded scenes and reflects the PTSNet has the ability to output pedestrian detection results quickly in the subway scene.
  \item [3)]
  In order to accurate measure the detection latency of the MetroNext-PST on the workstation and embedded platforms, the inference speed experiments are conducted in this paper, and the MetroNext-PST demonstrates faster inference speed, which means it is more suitable choice for speed priority applications on the embedded platforms.
\end{itemize}

The rest of this paper is structured as follows. Section II introduces the current research work on suppressing FPs in the pedestrian detection, followed by the explanation of the theory of the proposed PST algorithm in Section III, the experiment results of which are presented and discussed in Section IV. Section V is the conclusion.

\section{Related Work}
Given the rise of autonomous driving, pedestrian detection is expected to be key enabling and benefiting the automotive safety and driver assistance, where false positives in pedestrian detection results are a troublesome problem for it. Over the past decades, the main solutions can be divided into traditional methods and emerging deep learning-based methods, which briefly summarize in follow.

\subsection{The traditional methods}
The vision-based pedestrian detection methods obtain detection results through two key image processing steps: feature extraction and proposal classification. The traditional methods design various handcrafted features to extract pedestrian features and further processes them via following subnetwork to output detection results \cite{Ref5,Ref6,Ref7,Ref8,Ref9,Ref10}. Due to the weak feature extraction capability of the handcrafted filters, the extracted features are poorly semantic in foreground (pedestrian object) information and background (non-pedestrian object) information, making it difficult for the top classification network to distinguish between pedestrians and background images using these features, resulting in some false ones in the detection results. \cite{Ref11,Ref12,Ref13,Ref14,Ref15} have proposed different tricks to suppress these FPs, among them, Bertozzi et al. \cite{Ref12} creatively proposed an multi-resolution infrared vision pedestrian detection system, which designed a series of matched filters to avoid a number of FPs, and the efficient of this pedestrian detection system had been proved in various situations with lower false-positive rates. However, the traditional methods have limited learning ability and can't be adapted to the large intra-class variation of backgrounds and pedestrian, thus preventing further improvements in detection accuracy.

\subsection{The deep learning-based methods}
The significant accuracy gain has been witnessed on the ImageNet dataset by using CNN-based image processing methods \cite{Ref16}, which has motivated many researchers devote their efforts into this methods to solve the bottlenecks problem for a number of computer vision tasks \cite{Ref17, Ref18}. In the field of pedestrian detection, Szarvas et al. \cite{Ref19} first represented pedestrian attributes by using CNN and extracted pedestrian features, and then used these features to train SVM classifiers to recognize human objects from input images. Since then, numerous research works \cite{Ref20,Ref21,Ref22,Ref23,Ref24,Ref25,Ref26,Ref27,Ref28,Ref29,Ref30} continuously improved the pedestrian detection accuracy on popular benchmark datasets. For false positive detection results, an annoying and notorious problem, researchers have also proposed different technical routes to tackle this problem.

\begin{figure*}[]
	\centering
		\includegraphics[width=0.7\textwidth]{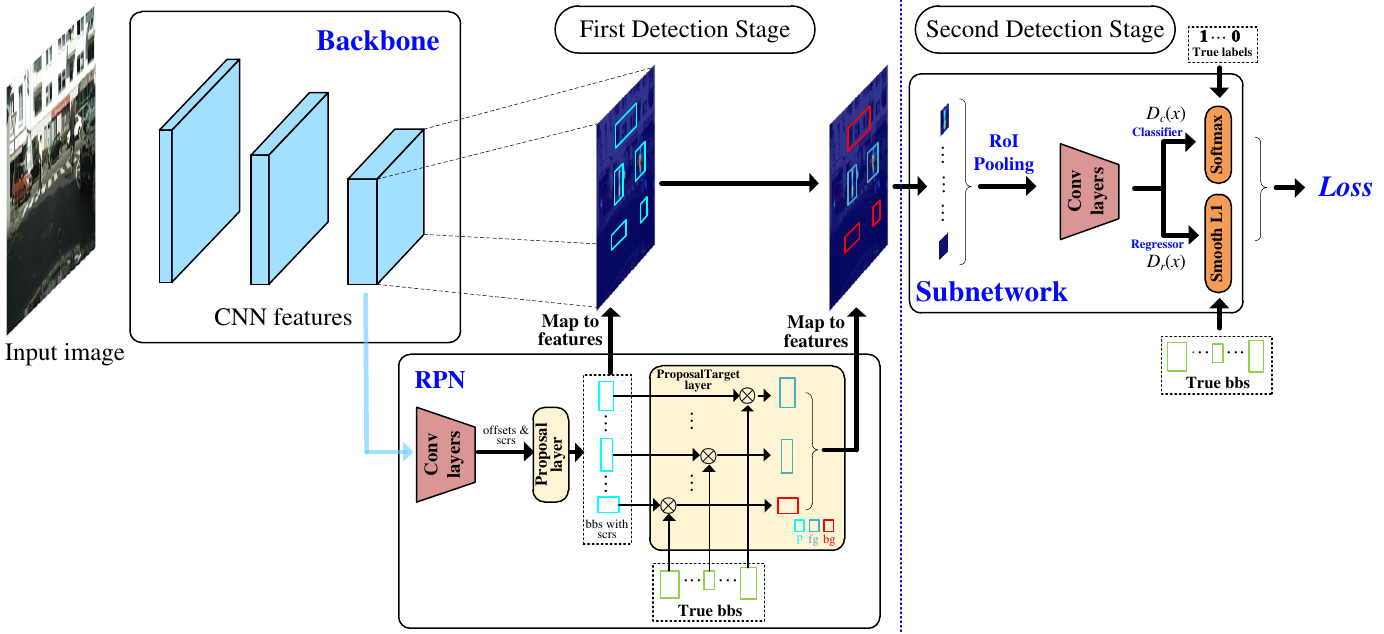}
    \caption{The training strategy of two-stage CNN-based pedestrian detection paradigm. Where ``offsets \& scrs" denotes the coordinate offsets and scores of predefined anchors. ``bbs" represents the bounding boxes. ``p" denotes the proposals, where foreground and background are represented by ``fg,bg". ``RoI Pooling" denotes the RoI Pooling layer. ``$D_c(x),D_r(x)$" denote the classifier and regressor of the subnetwork. $\bigotimes$ denotes the operator for calculating IoU values.}
	\label{FIG:1}
\end{figure*}

\textbf{3D convolution operation} Differ from 2D convolution operation, the 3D convolution operation can move in all 3-direction (height, width, channel of the feature maps) and, so it can directly process 3D perception data (e.g. Light Detection and Ranging data) and help the detector to describe the spatial relationships of objects in the 3D space. Gomez-Donoso et al. \cite{Ref31} proposed to combine 2D deep learning-based pedestrian detectors with a 3D CNN to wipe out pedestrian false ones, where the 2D deep learning-based pedestrian detectors is responsible to provide potential pedestrian proposals and 3D CNN is used to charge these proposals whether true positives. Similarly, Iftikhar et al. \cite{Ref32} adopted same detection pipeline to reject FPs and demonstrated that this method have better general capability in one-stage and two-stage CNN-based detection methods. The 3D convolution FPs suppression strategy additionally plugs a 3D network and uses it to process 3D LiDAR data to assist in determining whether the 2D detection pedestrian results are true ones or not. Unfortunately, this method will increase the computational complexity of the whole detection system, meanwhile, it can't be deployed in the application scenarios that are not equipped with 3D perception sensors, so the method doesn't have a good application prospect in the embedded edge scenarios.


\textbf{Multimodal information fusion} The multimodal information fusion reduces false positives on hard negative samples by efficiently fusing many different types of input data, while establishing recognition models and algorithms suitable for handling these fused data. Kolluri et al. \cite{Ref33} combines the multimodal object detector with Kernel Extreme Learning Machine (KELM) and Hybrid Salp Swarm Optimization (HSSO) algorithm to build IPDC-HMODL model, which adopts a image processing techniques to process two kind of input data: image frames and its ground truth images to help the IPDC-HMODL model to suppress the false ones in its pedestrian detection results. Kolluri et al. \cite{Ref34} proposed a CNN-based LiDAR-camera fusion mechanism and then design a neural network to confirm the pedestrian identification to decrease FPs. The comparison study have highlighted the enhanced outcomes of these methods. However, the integration of multimodal information faces serveral challenges. Firstly, the acquisition of multiple sensors can significantly increase its development costs. Secondly,  the alignment of information across various sensors requires additional processing steps, which promote the system's complexity. Thus, the advanced hardware to ensure the efficient and effective multimodal information fusion strategy.

\begin{figure*}[bp]
	\centering
		\includegraphics[width=0.9\textwidth]{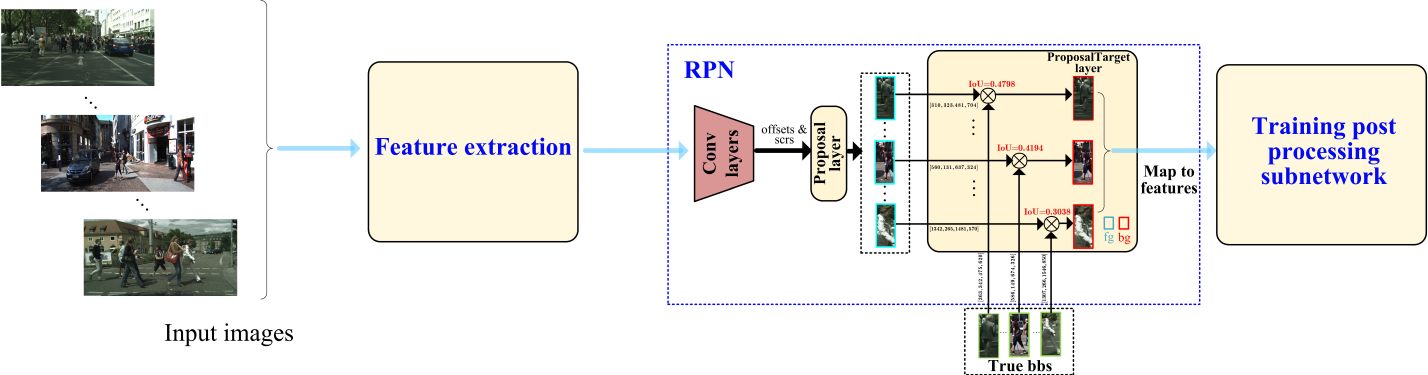}
    \caption{Some training samples are misclassified to negative ones using the IoU strategy.}
	\label{FIG:2}
\end{figure*}

\textbf{Non-Maximum Suppression strategy} A ideal NMS threshold for pedestrian detection is still a challenging, and the higher threshold brings more FPs while a lower one cause the higher miss rate. Liu et al. \cite{Ref35} proposed a adaptive NMS strategy that can compute suitable suppression threshold according to the object density, but the accurate object density estimation in complex real-life scenes is difficult task and this approach is useless when accurate density information is difficult to obtain. Unlike standard NMS strategy, Huang et al. \cite{Ref36} design a novel Representative Region NMS (R$^2$NMS) strategy to calculate Intersection over Union (IoU) of two objects, where representative region boxes of two objects are used to compute this value, and a Paired-Box Model (PBM) is proposed to responsible to predict pedestrian's representative region boxes. However, the problem with this method is how to ensure that the PBM has strong generalisation capabilities that can provide accurate predictions of representative region boxes of pedestrians in dynamically changing real-life scenarios, besides, it brings more human annotation work to label human visible body when establishing training datasets. \cite{Ref37, Ref38} had a similar technical route to optimise the NMS algorithm, and although the false positive examples in the pedestrian detector have been reduced to a certain extent, the aforementioned problem has still not been solved.

\textbf{Feature enhancement} The core of feature enhancement method construct various image processing module to strengthen and enrich pedestrian features to help the detector to recognize hard samples. \cite{Ref29} et al. adopted two key components: attention modules and reverse fusion blocks to build a semantic attention fusion mechanism to increase the discriminability of detector. \cite{Ref40} et al. designed a Pose-Embedding Network that combines human pose information with visual description, and used the pedestrian pose information to re-evaluate the confidence scores of pedestrian proposals and eliminate the false positives with high confidence scores. However, the main drawback of these methods is that the constructed feature information extraction module further increases the computational cost of the deep learning methods, making them more demanding on the computational resources of the target platforms, which hinders their widespread use on low-end computing devices.

In summary, although several research works have been carried out attempting to effectively remove pedestrian false positives, to the best of my knowledge, there is still a lack of a method that is suitable for deployment on embedded devices with good pedestrian false positives suppression with no extra computational cost, therefore in this paper, we have devoted our efforts to conduct a research work in this direction by proposing a training algorithm to help the two-stage CNN pedestrian detection method to remove pedestrian false ones.

\section{Methodology}
In this section, we first introduce the detection pipeline of the two-stage CNN detection method and its deficiency to solve the problem of pedestrian detection, then the theory of the PST algorithm is explained and how to use it to solve this problem.

\begin{figure*}[]
	\centering
		\includegraphics[scale=.7]{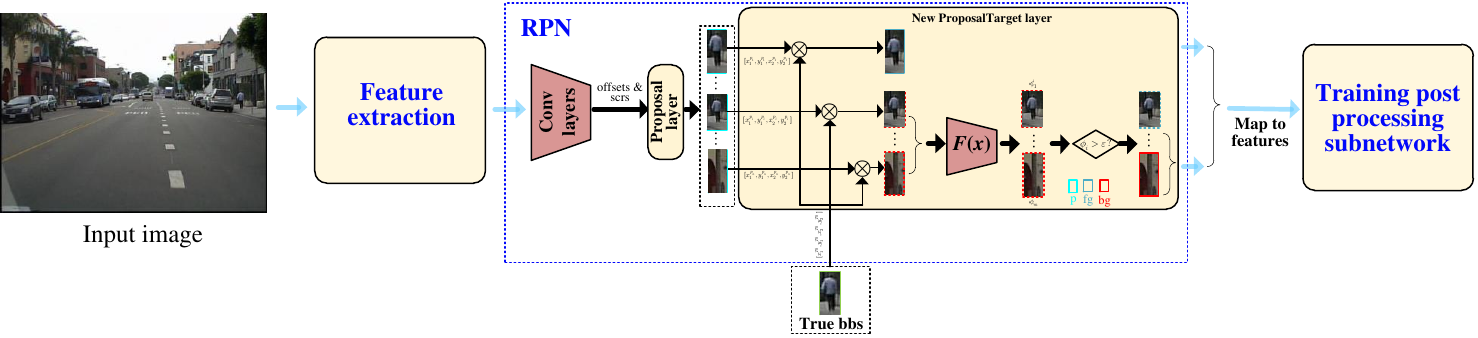}
    \caption{The processing pipeline of the PST algorithm, where $F(x)$ denotes the pedestrian sensitive classifier. $\phi_i$ represents the pedestrian confidence of the proposal. $\varepsilon$ represents the confidence threshold for pedestrians. All coloured dotted bounding boxes indicate these proposal are not used for training subnetwork.}
	\label{FIG:3}
\end{figure*}

\subsection{The training strategy of the two-stage CNN-based pedestrian detection paradigm}
As is shown in Fig.~\ref{FIG:1}, the two-stage CNN-based pedestrian detection paradigm consists the first detection stage and the second detection stage. The former is used to produces the pedestrian proposals and the latter processes these proposals to output final pedestrian detection results. The former has two key components: backbone network and Region Proposal Network (RPN). In training stage, the backbone network extract deep convolutional features from the input images, and then the several convolutional layers of RPN utilises these features to obtain coordinate offsets and scores of predefined anchors and send to the following the proposal layer to produce pedestrian proposals. These proposals contain some foreground ones and a lot of background ones, which needs to be accurately classified into positive and negative proposals to effectively guide the training process of the subnetwork. Where the proposal target layer is responsible for finishing this job, which adopts Intersection over Union (IoU) strategy to classify pedestrian proposals. However, the Iou strategy uses thresholds to classify pedestrian proposals, which will misclassify some ones containing human body information as negative ones (detailed analyses will be presented in the following subsections), and these misclassified proposals are mixed into the training samples, which will directly influence the training process of the subnetwork, reduce its pedestrian/background distinguishing ability, and inevitably lead to some FPs appearing in the pedestrian detection results.

\subsection{The drawback of IoU strategy}
IoU strategy is a common metric in the object detection and used to measure the match degree of two bounding boxes. The higher the IoU value of the two object, the better the match between them. Generally, an IoU threshold is predefined to evaluate the hit rate of the predictions to the ground truth. When the IoU value of them is larger than the predefined threshold, the predictions hit the GT (considered to the foreground), otherwise they don't hit it (considered to the background). Therefore, IoU threshold is a very important hyperparameter, and how to choose an appropriate IoU threshold has always been a tricky problem \cite{Ref38}. In addition, since the IoU threshold is pre-set and cannot be changed adaptively, it is not suitable for dealing with object detection problems with large scale variations such as pedestrian detection. This can be observed from its computational formula.
Given two bounding boxes $p=\{x_1^p, y_1^p, x_2^p, y_2^p\}$ and $tb=\{x_2^{tb}, y_1^{tb}, x_2^{tb}, y_2^{tb}\}$, their IoU value is computed
\begin{eqnarray}
IoU=\left| \frac{p\cap tb}{p\cup tb} \right|
\end{eqnarray}
where $\left| p\cap tb \right|$ and $\left| p\cup tb \right|$ represent the area of interaction and union region of $p$ and $tb$ respectively.
According to Equation (1), if the denominator is much larger than the numerator, a lower IoU score below the IoU threshold will be obtained and the corresponding proposal will be regarded as a negative training sample, which may occur in pedestrian detection. As shown in Fig.~\ref{FIG:2}, although several proposals generated by the RPN from the input images  contain human bodies, their IoU values after IoU computation with true bounding boxes are lower than the IoU threshold, and thus they are misclassified as negative samples, which directly affects the training process of the subnetworks so that two-stage CNN-based pedestrian detection paradigm can't effectively classify pedestrians and backgrounds, which leads to some FPs. A direct solution is to lower the IoU threshold so that these ones are correctly classified, leading to effective training of the network. However, the side effect of which is that a lower IoU threshold will bring some proposals with poorer localisation quality, affecting the localisation accuracy of the network. Therefore, during the training stage, only only adjusting the IoU threshold can't effectively solve the above problem, and it is necessary to use novel methods to increase the classification ability of the network between positive and negative training samples and to guide the effective training of the subnetwork. For this reason, this paper proposes the PST algorithm to meet this demand.

\begin{figure*}[bp]
	\centering
		\includegraphics[scale=.9]{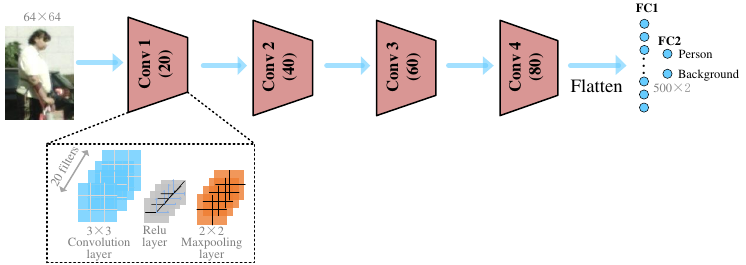}
    \caption{The architecture of the pedestrian-sensitive classifier. ``Conv" denotes its convolutional block and the number in the bracket means the number of convolutional filters. ``FC" stands for the Fully Connected layer.}
	\label{FIG:4}
\end{figure*}

\subsection{The PST algorithm}
As mentioned above, false positives occur because the IoU strategy misclassifies the training samples, thereby affecting the training process of the network and weakening its classification ability. Thus the core problem is to provide an accurate classification of positive and negative samples to the network and guide it to train effectively. During the training process of the two-stage pedestrian detection paradigm, the obtained network weights oriented to get accurate pedestrian localisation yet to accurate pedestrian classification is weak. Therefore, this paper adopts the cascading strategy to designs the PST algorithm to enhance the pedestrian/background classification capability of this paradigm , which construct an independent module to empower this paradigm with the pedestrians sensitivity and help it accurately differentiate pedestrians/backgrounds during the training process without the influence of weights update. The theory of the PST algorithm is described below.

\textbf{Overall processing pipeline} The paper proposes the PST algorithm and plugs it into the two-stage CNN-based pedestrian detection paradigm, which brings the step of evaluating the pedestrian information quantity for each proposal and changes the allocation process of positive and negative proposals in the proposaltarget layer. The overall processing pipeline of the PST algorithm is demonstrated in Fig.~\ref{FIG:3}, in which the input image $I$ is seed to the two-stage CNN-based pedestrian detection paradigm, and then a series of proposals $P$ are generated from the proposal layer, which can be denoted as
\begin{eqnarray}
\emph{\textbf{P}}=\{p_1,\cdots,p_n\},
\end{eqnarray}
where $i$th proposal is represented as $p_i=\{x_1^{p_i}, y_1^{p_i}, x_2^{p_i}, y_2^{p_i}\}$, $1\leq\emph{i}\leq\emph{n}$.
These proposals are computed with the true bounding box $tb$ to obtain their IoU scores.

\begin{eqnarray}
\emph{\textbf{S}}_{IoU}=\left\{ \left| \frac{p_1\cap tb}{p_1\cup tb} \right|,\cdots,\left| \frac{p_n\cap tb}{p_n\cup tb} \right| \right\}
\end{eqnarray}
where the IoU score of $i$th proposal is denoted as $IoU_i = \left| \frac{p_i\cap tb}{p_i\cup tb} \right|$. Comparing each IoU score with the threshold $\varepsilon_{IoU}$, the corresponding proposals can be divided into positive and negative training samples, as is

\begin{eqnarray}
  \begin{aligned}
{\emph{\textbf{P}}^{+}} &= \{p_i, IoU_i \geq \varepsilon_{IoU}, 1\leq\emph{i}\leq\emph{n}\} \\
{\emph{\textbf{P}}^{-}} &= \{p_j, IoU_j < \varepsilon_{IoU}, 1\leq\emph{j}\leq\emph{n}\}
  \end{aligned}
\end{eqnarray}

In the new proposaltarget layer, the ${\emph{\textbf{P}}^{+}}$ is treated as the true positive samples, but the ${\emph{\textbf{P}}^{-}}$ can't be seen as the true negative samples and are required a extra processing procedure to reevaluate the human information of each proposals, and the proposals with rich human body information will be reject in this step. To this end, a new-designed pedestrian-sensitive classifier is adopt to achieve this goal, which process the ${\emph{\textbf{P}}^{-}}$ corresponding input image slices (represented as $\textbf{I}_{\emph{\textbf{P}}^{-}}$) and output their human body information scores.

\begin{eqnarray}
\boldsymbol\Phi=F(\textbf{I}_{\emph{\textbf{P}}^{-}}),\ \ \boldsymbol\Phi=({{\phi}_{1}},...,{{\phi}_{m}})
\end{eqnarray}

When scores in the $\boldsymbol\Phi$ higher than the threshold, the corresponding proposals will be omitted from ${\emph{\textbf{P}}^{-}}$ , which solves the problem that these samples containing rich human torso information mislead the training process of the subnetwork. In addition, these proposals are not used as positive ones to train the subnetwork due to the fact that the positioning accuracy of these proposals is poor, which requires larger coordinate offsets and affects the convergence of the training process of the network. Thus, the new negative samples are

\begin{eqnarray}
{\emph{\textbf{P}}^{-}_{n}}=\{p^-_i, {\phi}_{i} < \varepsilon, 1\leq\emph{i}\leq\emph{m}\}
\end{eqnarray}

Finally, the ${\emph{\textbf{P}}^{+}}$ and ${{\emph{\textbf{P}}^{-}_{n}}}$ is merged to a new training proposal to guide the subnetwork effectively training to construct a stronger pedestrian/background classification capability.

\begin{eqnarray}
\emph{\textbf{P}}_t={\emph{\textbf{P}}^{+}} \cup {\emph{\textbf{P}}^{-}_{n}}
\end{eqnarray}

\textbf{Pedestrian-sensitive classifier} The key component of the PST algorithm is the pedestrian-sensitive classifier, which helps the PST algorithm accurately classify the pedestrian/background proposals. Moreover, the classifier requires to have low computational burden to not slow down the training speed of the paradigm in each training epoch. As a result, the newly designed classifier need to have a performance balance between stronger classification capabilities and less computational resources. To achieve this goal, this paper utilizes the following methods to construct the pedestrian-sensitive classifier.
The pedestrian-sensitive classifier is a CNN-based classifier, which can directly process the images and extract the image features to classify the object. The architecture of the classifier can be expressed as

\begin{eqnarray}
F(x) =(L_1 \circ,\cdots,\circ L_i \circ,\cdots,\circ L_k)(\emph{\textbf{I}})
\end{eqnarray}
where $L_i$ is a network layer and $k$ means that the classifier totaly has $k$ layers. $\circ$ denotes the layer connection.
Convolutional neural network has many types of network layers, this paper combines the convolutional layer, pooling layer plus fully connected layer to construct the classifier, in which the convolutional layer is used to extract the image features and the pooling layer is used to compress the features information, and the fully connected layer is adopted to map extracted features to the label space. Then, the $F(x)$ can be as
\begin{eqnarray}
F(x) = (L^c_1 \circ L^p_1 ,\cdots,\circ L^c_i \circ L^p_i \circ,\cdots,\circ L^{FC}_k)(\emph{\textbf{I}})
\end{eqnarray}
where $L^c_i,L^p_i$ denote $i$th convolutional or pooling layer respectively. $L^{FC}$ stands for the last fully connected layer. Let $\emph{\textbf{W}}^c_i, \emph{\textbf{W}}^p_i,i=1,2,\cdots ,k$ represent sampling matrix of convolutional and pooling layers, and $\emph{\textbf{W}}^{FC}$ stand for the fully connected weights. $\textbf{\emph{X}}\in {\textbf{\emph{R}}^{H\times W\times C}}$ denotes input features of different layers. The equation 9 can be rewritten as

\begin{equation}
    \begin{aligned}[b]
F(x) &= \{\{{\textbf{\emph{W}}^c_{k-1}} \ast \{ \cdots \{{\textbf{\emph{W}}^c_{2}}*[({\textbf{\emph{W}}^c_{1}}*\textbf{\emph{I}})\downarrow {\textbf{W}^p_{1}}]\}\downarrow {\textbf{W}^p_{2}}\cdots \}\} \\
& \downarrow {\textbf{W}^p_{k-1}}\}\emph{\textbf{W}}^{FC}
    \end{aligned}
\end{equation}
where ``$\ast,\downarrow$" means the convolution and pooling operation. \cite{Ref50} has pointed out the computation complexity of CNN is determined by the convolution operation and its total number of network parameters is mainly occupied by the fully connected layer. Therefore, this paper controls the computation complexity of the convolution layer to reduce the computational cost of the classifier, which is controlled by the following equations
\begin{eqnarray}
{\rm{M}_{Conv}} = {f^2}C_{i}C_{f}{W}'{H}'
\end{eqnarray}
\begin{eqnarray}
    \begin{aligned}
{H}' = & [(H-f+2p)/s]+1 \\
{W}' = & [(W-f+2p)/s]+1
    \end{aligned}
\end{eqnarray}
where $\rm{M}_{Conv}$ stands for the computation complexity of the convolution layer. $f$ is the size of the filter. $C_{i},C_{f}$ represents the number of channels in the input image and convolution filters. $p,s$ are the padding and stride parameters, respectively. According to Eqs. (11)-(12), the computational complex of the convolution layer is controlled by $H,W,p,s,f,C_{i},C_{f}$, among which only the $C_{f}$ has optimize selection space, so this paper need to optimise $f,C_{i},C_{f}$ to reduce the computational complexity of the classifier.

For the fully connected layer, its total number of network parameters is computed by
\begin{eqnarray}
\rm{N}_{FC} = IJ
\end{eqnarray}
where $\rm{N}_{FC}$ stands for total number of network parameters of the fully connected layer, and $I,J$ are its input and output vectors' size. Since $I$ is determined by the feature map shape of previous layers, so this paper can decreases $J$ to reduce the total parameters of the pedestrian-sensitive classifier.

Finally, the total number of layers of the classifier needs to be determined to ensure that its top layer has a receptive field to effectively cover the input image. The equation for calculating the receptive field of CNN is as follows \cite{Ref40}.

\begin{eqnarray}
{{r}_{n}}={{r}_{n-1}}+({f}_{n}-1)\prod\limits_{i=1}^{n-1}{{{s}_{i}}}
\end{eqnarray}
where $r_n, r_{n-1}$ represent the receptive field of $n$th layer and $(n-1)$th layer, respectively.

Using the equation (14), the receptive field of the top layer of the 9-layer CNN covers more than a quarter of a 64$\times$64 image, which helps the classifier to effectively extract human features from the input image.
In summary, this paper utilizes the following strategies to the pedestrian-sensitive classifier, the structure of which is shown in Fig.~\ref{FIG:4}.
\begin{itemize}
  \item [1)]
  This paper adopts $3\times3$ convolution filter and the shorter output vector length of fully connected weights to cut down the computation burden of the classifier.
  \item [2)]
  The 9-layer network structure ensures that the top layer of the classifier has a ressonable receptive field to cover the input image.
\end{itemize}

To demonstrate the PST algorithm in detail, its pseudo-code is listed in the Algorithm 1.

\begin{table}[!h]
\centering
\begin{tabular}{l}
\hlinewd{1pt}
\textbf{Algorithm 1 PST Algorithm}                                                                                                                                                                                                                                                                                                                                                                                                                                                                                                                                           \\ \hline
\begin{tabular}[c]{@{}l@{}} \textbf{procedure} \textbf{PST}(Input crowd image: $\textbf{I}$) \\ //Initialize, define the proposal set $\textbf{P}$, the IoU scores set $\textbf{S}_{IoU}$, positive \\and
negative training samples ${\emph{\textbf{P}}^{+}}$, ${{\emph{\textbf{P}}^{-}}}$, pedestrian confidence scores for \\ image slices set $\boldsymbol\Phi$, the refined negative training samples set ${{\emph{\textbf{P}}^{-}_{n}}}$ and \\ the refined training sample set $\emph{\textbf{P}}_t$. Give the confidence score threshold $\varepsilon$.\\ 1: \textbf{init} $\textbf{P},\textbf{S}_{IoU},\emph{\textbf{P}}^{+},{\emph{\textbf{P}}^{-}},\boldsymbol\Phi,{\emph{\textbf{P}}^{-}_{n}},\emph{\textbf{P}}_t$, $\varepsilon$\\ 2. Obtain feature maps $\textbf{X}_{\rm{backbone}}$ by processing $\textbf{I}$.\\3. Obtain proposals $\emph{\textbf{P}}$ on $\textbf{X}_{\rm{backbone}}$.\\4. Obtain ${\emph{\textbf{P}}^{+}}$, ${{\emph{\textbf{P}}^{-}}}$ using the IoU algorithm. \\\kern 0.8pc4.1 \textbf{for} $i\leftarrow 1$ \emph{to} $i\leftarrow n$ \textbf{do}\\\kern 0.8pc4.2 \kern 0.6pc $IoU_i = \left| \frac{p_i\cap tb}{p_i\cup tb} \right|$\\\kern 0.8pc4.3 \kern 0.6pc \textbf{if} $IoU_i \geq \varepsilon_{IoU}$ ${\emph{\textbf{P}}^{+}} \cap \{p_i\}$ \textbf{else} ${\emph{\textbf{P}}^{-}} \cap \{p_i\}$\\5. Obtain $\textbf{I}_{\emph{\textbf{P}}^{-}}$ based on the coordinates of $\emph{\textbf{P}}^{-}$\\6. Obtain refined negative proposals ${\emph{\textbf{P}}^{-}_{n}}$ using the classifier $F(x)$ \\\kern 0.8pcto process $\textbf{I}_{\emph{\textbf{P}}^{-}}$\\\kern 0.8pc6.1 \textbf{for} $i\leftarrow 1$ \emph{to} $i\leftarrow m$ \textbf{do}\\\kern 0.8pc6.2 \kern 0.6pc ${\phi}_i=F(\textbf{I}_{p_i^{-}})$\\\kern 0.8pc6.3 \kern 0.6pc \textbf{if} ${\phi}_i < \varepsilon$ ${\emph{\textbf{P}}^{-}_{n}} \cap \{p_i\}$\\8. Obtain refined training proposals $\emph{\textbf{P}}_t={\emph{\textbf{P}}^{+}} \cup {\emph{\textbf{P}}^{-}_{n}}$ to guide training \\\kern 0.8pc process of the subnetwork.
\end{tabular} \\ \hlinewd{1pt}
\end{tabular}
\end{table}

\section{Experiments and results}
\subsection{System Setting}
\emph{Experimental platform}. The evaluation experiments of the proposed methods have been done on both a workstation and an embedded platform, whose specifications are outlined in Table I. Of significance, the embedded development board utilized closely mirrors the resource constraints typical of metro surveillance systems. This alignment ensures that the study effectively demonstrates the established detector's capability for accurate pedestrian detection within the constraints of resource-limited metro environments, validating its practical feasibility.

\begin{table}[]
	\centering
	\caption{The detail hardware and software information of experiment platforms}\label{Tbl3}
	\setlength{\tabcolsep}{0.2em}
	\begin{tabular}{lll}
		\hline
		\multirow{2}{*}{\textbf{\begin{tabular}[c]{@{}l@{}}Software \& \\ Hardware\end{tabular}}} & \multicolumn{2}{c}{\textbf{Platforms}}                \\ \cline{2-3}
		& Workstation           & Jetson nano   \\ \hline
		CPU                                                                                       & Intel Core i7-6950x   & ARM Cortex-A53 \\
		MEMORY                                                                                    & 64G                   & 4G                           \\
		GPU                                                                                       & NVIDIA TITAN X Pascal & NVIDIA Maxwell                \\
		Operating System                                                                           & \multicolumn{2}{c}{Ubuntu 18.04}                     \\ \hline
	\end{tabular}
\end{table}

\emph{Evaluation metrics}. This paper employs total parameters, inference time, and miss rate as key metrics in forthcoming experiments. These evaluation metrics provide a holistic assessment of memory usage, computational complexity, and detection accuracy of all detector in pedestrian detection. By emphasizing these aspects, the strengths of the PST algorithm can be underscore in enhancing pedestrian detection capabilities without incurring additional computational overhead, and also to reflect the the strengths and weaknesses of all detectors in different platforms. In addition, all experimental setups adopt an Intersection over Union (IoU) threshold of 0.5 as the default value.

\begin{table}[bp]
	\centering
	\caption{Results of Citypersons dataset from FasterRCNN using the PST algorithm}\label{Tbl1}
	\setlength{\tabcolsep}{0.2em}
	\begin{tabular}{ccccc}
		\hline
		\textbf{\begin{tabular}[c]{@{}l@{}}Detector @ \\ CityPersons\end{tabular}} & Backbone & MR(\%)       & GPU(ms/frame) & Params(M)    \\ \hline
		FasterRCNN-PST                                                                 & VGG16    & 73.63(-0.8) & 72(+0)      & 137.1(+0) \\
		\hline
	\end{tabular}
\end{table}

\begin{table*}[bp]
	\centering
	\caption{The performance of MetroNext-PST and the state-of-the-art detectors on the benchmark datasets}\label{Tbl3}
	\setlength{\tabcolsep}{0.2em}
	\begin{tabular}{llllllll}
		\hline
		\multirow{2}{*}{\textbf{Detectors}} & \multirow{2}{*}{\begin{tabular}[c]{@{}l@{}}Params\\ (M)\end{tabular}} & \multicolumn{2}{l}{\textbf{CUHK-Occ}}                                                                        & \multicolumn{2}{l}{\textbf{Caltech}}                                                                         & \multicolumn{2}{l}{\textbf{Citypersons}}                                                                     \\ \cline{3-8}
		&                                                                       & \begin{tabular}[c]{@{}l@{}}MR\\ (\%)\end{tabular} & \begin{tabular}[c]{@{}l@{}}GPU\\ (ms/frame)\end{tabular} & \begin{tabular}[c]{@{}l@{}}MR\\ (\%)\end{tabular} & \begin{tabular}[c]{@{}l@{}}GPU\\ (ms/frame)\end{tabular} & \begin{tabular}[c]{@{}l@{}}MR\\ (\%)\end{tabular} & \begin{tabular}[c]{@{}l@{}}GPU\\ (ms/frame)\end{tabular} \\ \hline
		Swin Transformer                                 & 45.31                                                                 & 24.80                                             & 105&                                                  58.72 & 106                                                      & 60.22                                                  & 106                                                      \\
		FPN                                 & 42.12                                                                 & 28.01                                             & 182&                                                  59.68 & 180                                                      & 64.28                                                  & 181                                                      \\
		FRCNN VGG16                         & 136.69                                                                & 36.59                                             & 70                                                       & 64.77                                             & 69                                                       & 73.63                                             & 72                                                       \\
		SSD512                              & 23.75                                                                 & 35.81                                             & 50                                                       & 69.26                                             & 50                                                       & 79.41                                             & 51                                                       \\
		Tiny YOLOV3                         & 8.66                                                                  & 53.33                                             & 3                                                        & 77.63                                             & 3                                                        & 86.82                                              & 3                                                        \\
		Pelee                               & 5.29                                                                  & 41.28                                             & 16                                                       & 74.42                                             & 16                                                       & 83.96                                              & 16                                                       \\
		EfficientDet                               & 3.90                                                                  & 40.62                                             & 32                                                       & 71.49                                             & 32                                                       & 80.49                                              & 33                                                       \\
		MetroNext                           & 4.56                                                                  & 37.81                                             & 25                                                       & 64.82                                             & 25                                                       & 71.87                                              & 26                                                       \\
		MetroNext-PST                           & 4.56                                                                  & 37.00                                             & 25                                                       & 64.63                                             & 25                                                       & 69.56                                              & 26                                                       \\
		\hline
	\end{tabular}
\end{table*}

\subsection{Datasets}
This paper utilizes the benchmark datasets and a metro scene dataset: SY\_Metro \cite{Ref4} to validate the feasibility of the proposed methods. The benchmark datasets include Caltech \cite{Ref43}, CUHK\_Occ \cite{Ref44} and CityPersons datasets \cite{Ref45} , which have been widely adopted in research works and can be used to reflect the effectiveness of the PST algorithm to solve the pedestrian detection, demonstrating the comprehensive performance competitiveness of the plain detector aided by the PST algorithm in hunting persons. The metro scene dataset is used in the pedestrian detection experiments in this paper because the metro scene is the main application of the proposed method, to achieve fast and accurate underground passenger detection results in this scene, and also because the benchmark datasets lack metro scenes, this dataset can further validate the generalization ability of the proposed method for pedestrian detection in cross scenes.

In this paper, the SY\_Metro dataset is randomly split into 50\% training dataset and 50\% testing dataset in the following experiments and the division of training and test data in the benchmark datasets can be found in the corresponding references. Additional details on these datasets can be learned in the references.

\subsection{Baseline detectors}
Extensive popular baseline detectors, including FasterRCNN \cite{Ref17}, SSD \cite{Ref46}, Tiny YOLOV3 \cite{Ref47}, FPN \cite{Ref48}, Pelee \cite{Ref49}, EfficientDet \cite{Ref50} and Swin Transformer \cite{Ref51} are employed for comprehensive comparison with the proposed method in pedestrian detecion, enabling a thorough validation of the PST algorithm's capability to enhance the detection accuracy of the plain detection network without extra computional cost.  By comparing with these established state-of-the-art detectors, the merits and demerits of detection performance of the PST-augmented detector can be witnessed.

\subsection{The ablation experiment}
To demonstrate the proficiency of the PST algorithm in enhancing the pedestrian detection accuracy of two-stage CNN-based detectors, the widely-adopted Faster R-CNN (FRCNN) as the backbone detector and evaluated it on the CityPersons dataset. The detailed experimental results are summarized in Table II, where ``MR" signifies the log average miss rate (a lower value indicating better performance), ``Params" represents the total number of parameters (``M" indicating millions), and ``GPU" refers to the inference duration per frame in milliseconds/frame (ms/frame). Variations relative to the plain detector are indicated in parentheses.
The CityPersons dataset's experiment results, as illustrated in Table II, exhibit a diminution in miss rate by 0.8\% accomplished without any extra computational expense. This improvement in detection accuracy of the plain detector for pedestrian detection without any additional computational burdens, which demonstrate the effectiveness of the PST algorithm in promoting the detector's accuracy while maintaining computational efficiency.

\subsection{Benchmark experiments}
To further validate the PST algorithm's effectiveness and versatility in improving pedestrian detection accuracy across diverse real-world scenarios, especially when applied to a compact detection network. This study adopts the compact yet powerful two-stage pedestrian detector: MetroNext, as the baseline detector. Its performance enhancement is fully evaluated against other widely-used detectors across benchmark datasets. A summary of the experimental results is presented in Table III, offering insights into the algorithm's capability in promoting the detector's overall detection performance under realistic conditions. In the Table III, the \textbf{MetroNext w/o} and \textbf{MetroNext w/} indicate that it is available without or with the PST algorithm. From the data in this table you can see that:

\begin{itemize}
  \item [1)]
  Empowered by the PST algorithm, the MetroNext witnesses a remarkable boost in its ability to hunt pedestrians across various scales within benchmark datasets. Specifically, in the CityPersons dataset, a detection accuracy improvement of 2.31\% is observed, highlighting the algorithm's capability in guiding the detector during training to differentiate between true positives and false detections. This enhancement in accuracy demonstrates the feasibility of incorporating our proposed PST algorithm into its learning phase of CNN-based pedestrian detection network, thereby can being a practical strategy tailored for CNN-based pedestrian detection networks in solving the pedestrian detection tasks.
  \item [2)]
  The MetroNext-PST variant has demonstrated outstanding detection capabilities, showing competitive performance compared to  other state-of-the-art detectors. Specially, despite the MetroNext has tiny model size and the fewer number of network parameters, it has better MR results within benchmark datasets, achieving 37.81\%, 64.82\%, and 71.87\% respectively. The integration of PST algorithm further boost the MetroNext detection accuracy, reducing the miss rate by up to 2\% compared to the original model without extra increase in total parameters and inference time. This means that the PST algorithm is effective to improve the model's detection accuracy, even for the small detection network.
  \item [3)]
  In contrast to established networks and other deep CNN, while they may excel in pedestrian detection accuracy, they also suffer from two significant drawbacks: a large number of network parameters and relatively slow inference speed. For instance, the Swin Transformer, despite with better accuracy, has a larger number of parameter of 45.31M and an inference speed of approximately 100 ms/frame, the overhead of which poses a challenge for deployment in resource-constrained embedded platforms. they also have drawbacks in huge number of network parameters and slower inference speed, which make these classifiers are not suitable to deploy in the embedded platform.  Meanwhile, comparing to the EfficientDet that are specifically oriented to the embedded platform, and MetroNext-PST strikes an optimal balance between accuracy, total parameter and inference speed. Thus, the proposed network is more suitable for embedded environment.
  \item [4)]
  Evaluating the detectors on the Caltech and CityPersons datasets, the \textbf{\emph{all}} settings mode provided by these datasets was utilized. In this mode, pedestrians are heavily occluded, posing a big challenge for the most powerful detection models. all detectors have higher MR results compared to their performance on the CUHK-Occ dataset. However, the established MetroNext-PST achieves better detection accuracy, thereby demonstrating  the ability of the PST algorithm to generalize and enhance the detection accuracy of small pedestrian detectors in complex real-world urban environments.
\end{itemize}

\textbf{Discussion} All detectors undergo training and evaluation using a unified data split. The proposed methods achieve performance gains on benchmark datasets, showcasing the PST algorithm's robustness in effectively guide the learning phase of plain detector to distinguish pedestrians and backgrounds. Aided by the PST algorithm, The established MetroNext-PST excels in terms of total parameters and inference speed yet lags behind larger models in accuracy, which is due to the smaller model scale of the plain detector plus the small pedestrian-sensitive classifier in the PST algorithm, which can't store sufficient semantic features to aid the detector in accurately recognizing pedestrians. Consequently, MetroNext-PST emerges as a competitive choice for pedestrian detection in resource-constrained edge devices. Future research could delve into the correlation between model size and pedestrian detection performance, and explore the boundary problem of the model size of the pedestrian-sensitive classifier required to effectively guide the learning process of pedestrian detection networks.

\begin{figure}[!h]
	\centering
	\subfigure[CUHK-Occ dataset]{
		\includegraphics[scale=.45]{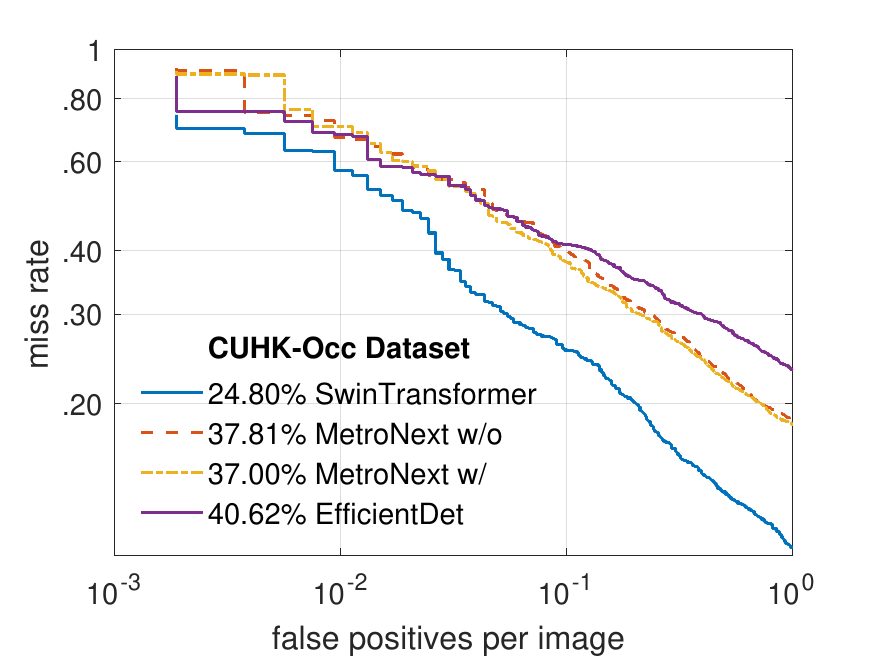}
	}
	\subfigure[Caltech dataset]{
		\includegraphics[scale=.45]{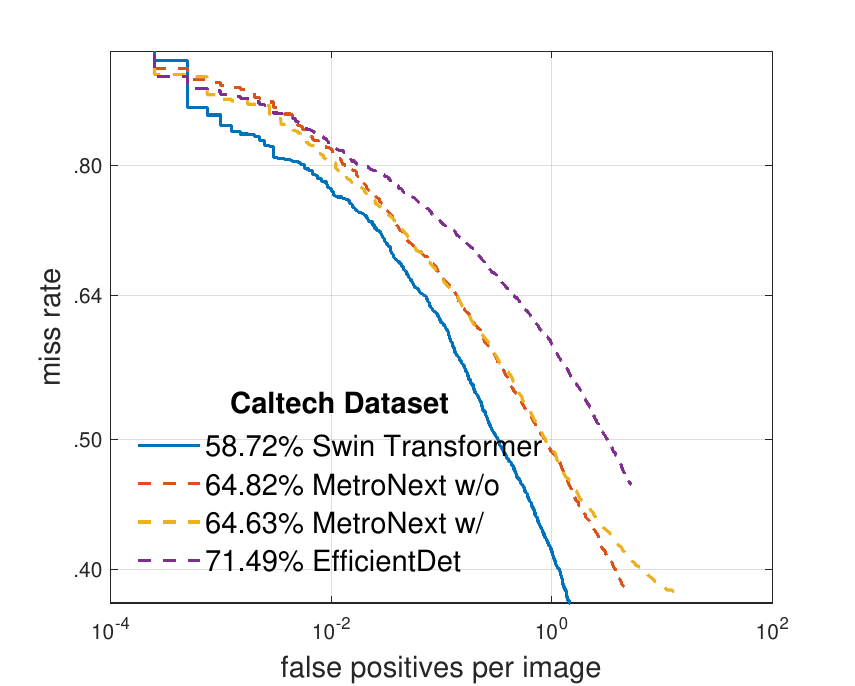}
	}
	\subfigure[CityPersons dataset]{
		\includegraphics[scale=.45]{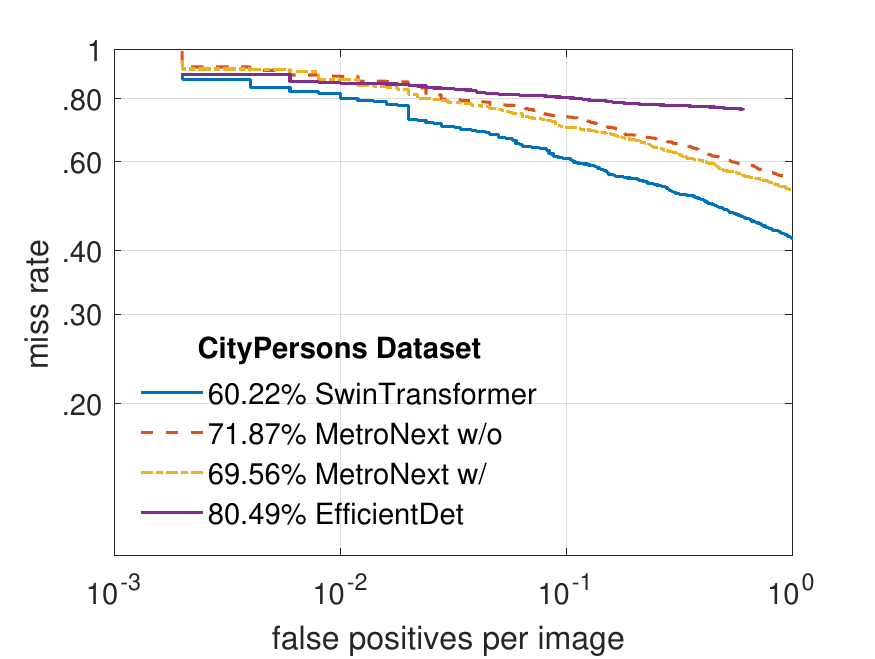}
	}
	\caption{Comparison to the state-of-the-art detectors on benchmark datasets. (a) CUHK-Occ dataset. (b) Caltech dataset. (c) CityPersons dataset.}
	\label{FIG:5}
\end{figure}

Figure 5 depicts the graphical representation of miss rates against false positives per images for Swin Transformer, MetroNext, MetroNext-PST and EfficientDet across benchmark datasets. These curves span miss rate intervals from 0.1 to 1 (adjusted to 0.4 to 1 for the Caltech dataset) and false positives per image ranging from $10^{-3}$ to $10^0$ (or from $10^{-4}$ to $10^2$ for the Caltech dataset). A lower curve signifies superior detection performance, and the legend accompanying the figure indicates the log average miss rates of these detectors.

\begin{table}[!h]
	\centering
	\caption{The performance of MetroNext-PST and the state-of-the-art detectors on SY-Metro datasets}\label{Tbl3}
	\setlength{\tabcolsep}{0.2em}
	\begin{tabular}{llllll}
		\hline
		\multirow{2}{*}{\textbf{Detectors}} & \multirow{2}{*}{\begin{tabular}[c]{@{}l@{}}Params  \\ (M)\end{tabular}} & \multicolumn{2}{l}{\textbf{SY-Metro}}                                                                           &                                                                     \\ \cline{3-6}
		&                                                                         & \begin{tabular}[c]{@{}l@{}}MR \\ (\%)\end{tabular} & \begin{tabular}[c]{@{}l@{}}GPU  \\ (ms/frame)\end{tabular} &  &  \\ \hline
		Swin Transformer                                 & 45.31                                                                   & 18.11                                              & 106                                                                                                            \\
		FPN                                 & 42.12                                                                   & 21.33                                              & 183                                                                                                               \\
		FRCNN VGG16                         & 136.69                                                                  & 29.99                                              & 71                                                                                                                  \\
		SSD512                              & 23.75                                                                   & 26.16                                              & 51                                                                                                                  \\
		Tiny YOLOV3                         & 8.66                                                                    & 42.29                                              & 3                                                                                                                  \\
		Pelee                               & 5.29                                                                    & 32.25                                              & 16                                                                                                                  \\
		EfficientDet                               & 3.90                                                                    & 27.27 & 32                                                         \\
		MetroNext                           & 4.56                                                                    & 26.70                                              & 26                                                         \\
		MetroNext-PST                       & 4.56                                                                    & 24.68                                              & 26                                                                                                                  \\
		\hline
	\end{tabular}
\end{table}

\subsection{The SY-Metro experiment}
In pursuit of validating the PST algorithm's universal effectiveness and exploring its capacity to guide compact detection models in distinguishing pedestrians from background across various real-life scenarios. This experiment employs the SY-Metro dataset that encapsulates the peculiarities of metro scenes, thereby extending the algorithm's validation into a distinctive, high-density urban context. Utilizing the SY-Metro dataset, The MetroNext-PST and other state-of-the-art detectors are validated on this dataset and the detailed experimental results are summarized in Table IV, which reveal several insights:

\begin{itemize}
	\item [1)]
	With the help of the PST algorithm, the MetroNext-PST achieves competitive detection performance. Apart from the Swin Transformer and FPN, MetroNext-PST boasts a notably lower miss rate of 24.68\% with a few parameters and considerable inference time. This achievement further supports the PST algorithm's versatility in boosting the pedestrian detection accuracy of the plain detector. In addition, given the limited hardware resources of the embedded system for deploying online passenger detectors in metro station, the comprehensive performance of the MetroNext-PST becomes a more viable option compared to other detectors.
	\item [2)]
	Metro stations are settled scenes compared to complicated outdoor scenes in other benchmark datasets and MetroNext has achieved low miss rate. Leveraging this foundation, the integration of the PST algorithm into the training process effectively guides the network learning process, enabling it to develop an accurate pedestrian classification capability.and therefore further improves the pedestrian detection accuracy of the base detection model (up to 2\% MR value reduction), demonstrating the reliability of the PTS algorithm in steadily improving the detection performance of the base network model.
\end{itemize}

Fig. 6 draws the miss rate versus false positives per image of all models on the SY\_METRO dataset, which clearly illustrates the detection prowess of each model in identifying metro passengers. As is shown in Fig. 6, our model achieves a competitive metro passenger detection ability compared to other competitors.

\begin{figure}[!h]
	\centering
	   \includegraphics[scale=.45]{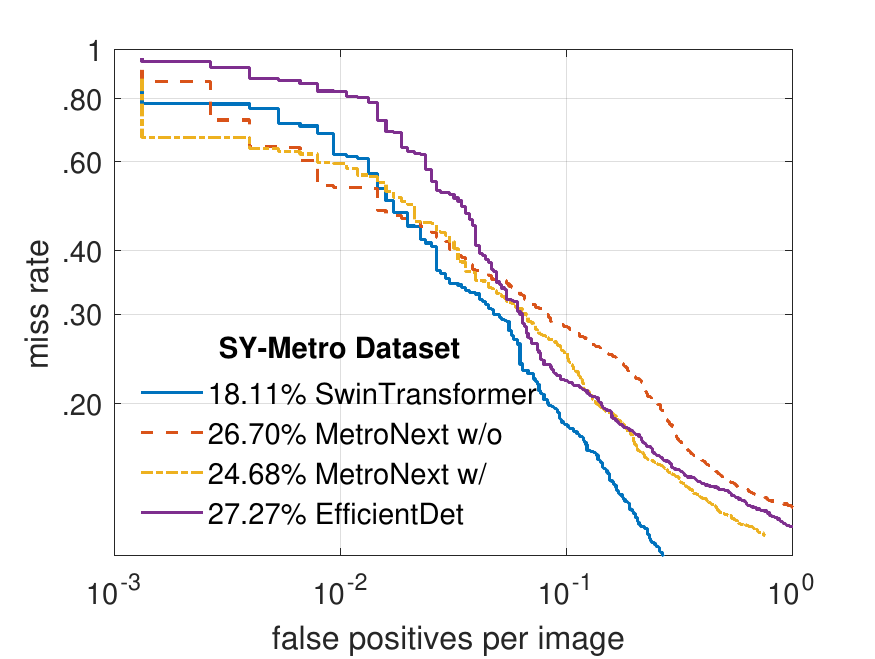}
    \caption{Comparison to the state-of-the-art detectors on SY-Metro datasets}
	\label{FIG:6}
\end{figure}

%
%

\subsection{The experiment on the embedded development board}
The experiments on the workstation have shown that the proposed classifier has the potential to be deployed on the embedded platforms. In order to accurately estimate its actual inference speed on the embedded development board with limited hardware resources, this paper adopts the C++ language write our model's forward inference program for metro crowd density estimation, and then test the inference speed of this program on the SY-Metro dataset. The specific experimental results are shown in Table VI, showing that:

\begin{itemize}
  \item [1)]
  Our model demonstrates a fast inference speed of 358ms when processing a image, which means that it can deliver the pedestrian detection results in the carriage to the metro video surveillance system quickly.
  \item [2)]
  The experimental results only show the inference speed of the proposed detector on an ARM CPU, which will be further enhanced when it is placed on a more powerful hardware device (e.g., ARM GPU).
\end{itemize}

\begin{table}[!h]
\centering
\caption{The inference speed of our model and the state-of-the-art classifiers on Jetson Nano.}\label{Tbl4}
\setlength{\tabcolsep}{0.2em}
\begin{tabular}{llllll}\hline
\textbf{Models}             & \textbf{MetroNext-PST} &  &  &  &  \\\hline
$\rm{Speed_{CPU}}$(ms/frame) & 358    &        &      &        & \\ \hline
\end{tabular}
\end{table}


\section{Conclusion}
In this paper, an elaborate pedestrian-sensitive training algorithm named the PST algorithm has been proposed to improve the pedestrian detection accuracy by removing FPs. The ablation experiment has demonstrated that the proposed algorithm is effective and practicable to promote the detection performance of mainstream detection network on the popular pedestrian benchmark dataset. Combing the PST algorithm with MetroNext, the MetroNext-PST is established and then validated on various pedestrian benchmark datasets. The experiment results have demonstrated that the MetroNext-PST is a more competitive pedestrian detector, and the PST algorithm can efficiently remove the FPs generating from the plain MetroNext. In addition, the experiments ofinference speed have supported that the MetroNext-PST has the potential to achieve better balance on accuracy, power and speed on embedded vision hardware. In summary, the PST algorithm can efficiently improve the model's detection accuracy without extra computation burden, and MetroNext-PST is a more practical pedestrian detector, especially for embedded vision tasks.

\ifCLASSOPTIONcaptionsoff
  \newpage
\fi

\bibliographystyle{splncs03}
\bibliography{example}

\begin{thebibliography}{10}
\providecommand{\url}[1]{\texttt{#1}}
\providecommand{\urlprefix}{URL }

\bibitem{Ref1}
Brunetti, A., Buongiorno, D., Trotta, G.F., Bevilacqua, V.: Computer vision and
  deep learning techniques for pedestrian detection and tracking: A survey.
  Neurocomputing  300,  17--33 (2018)

\bibitem{Ref2}
Benenson, R., Omran, M., Hosang, J., Schiele, B.: Ten years of pedestrian
  detection, what have we learned? In: Computer Vision-ECCV 2014 Workshops:
  Zurich, Switzerland, September 6-7 and 12, 2014, Proceedings, Part II 13. pp.
  613--627. Springer (2015)

\bibitem{Ref3}
Cao, J., Pang, Y., Xie, J., Khan, F.S., Shao, L.: From handcrafted to deep
  features for pedestrian detection: A survey. IEEE Transactions on Pattern
  Analysis and Machine Intelligence  44(9),  4913--4934 (2022)

\bibitem{Ref4}
Liu, Q., Guo, Q., Wang, W., Zhang, Y., Kang, Q.: An automatic detection
  algorithm of metro passenger boarding and alighting based on deep learning
  and optical flow. IEEE Transactions on Instrumentation and Measurement  70,
  1--13 (2021)

\bibitem{Ref5}
Oren, M., Papageorgiou, C., Sinha, P., Osuna, E., Poggio, T.: Pedestrian
  detection using wavelet templates. In: Proceedings of IEEE computer society
  Conference on computer vision and pattern recognition. pp. 193--199. IEEE
  (1997)

\bibitem{Ref6}
Viola, Snow: Detecting pedestrians using patterns of motion and appearance. In:
  Proceedings ninth IEEE international conference on computer vision. pp.
  734--741. IEEE (2003)

\bibitem{Ref7}
Dalal, N., Triggs, B.: Histograms of oriented gradients for human detection.
  In: 2005 IEEE computer society conference on computer vision and pattern
  recognition (CVPR'05). vol.~1, pp. 886--893. Ieee (2005)

\bibitem{Ref8}
Wang, X., Han, T.X., Yan, S.: An hog-lbp human detector with partial occlusion
  handling. In: 2009 IEEE 12th international conference on computer vision. pp.
  32--39. IEEE (2009)

\bibitem{Ref9}
Doll{\'a}r, P., Tu, Z., Perona, P., Belongie, S.J.: Integral channel features.
  In: Bmvc. vol.~2, p.~5. London, UK (2009)

\bibitem{Ref10}
Schwartz, W.R., Kembhavi, A., Harwood, D., Davis, L.S.: Human detection using
  partial least squares analysis. In: 2009 IEEE 12th international conference
  on computer vision. pp. 24--31. IEEE (2009)

\bibitem{Ref11}
Bertozzi, M., Broggi, A., Fascioli, A., Graf, T., Meinecke, M.M.: Pedestrian
  detection for driver assistance using multiresolution infrared vision. IEEE
  transactions on vehicular technology  53(6),  1666--1678 (2004)

\bibitem{Ref12}
Bertozzi, M., Broggi, A., Lasagni, A., Rose, M.: Infrared stereo vision-based
  pedestrian detection. In: IEEE Proceedings. Intelligent Vehicles Symposium,
  2005. pp. 24--29. IEEE (2005)

\bibitem{Ref13}
Nam, W., Doll\'{a}r, P., Han, J.H.: Local decorrelation for improved pedestrian
  detection. In: Proceedings of the 27th International Conference on Neural
  Information Processing Systems - Volume 1. p. 424–432. MIT Press,
  Cambridge, MA, USA (2014)

\bibitem{Ref14}
Suzuki, Y., Deguchi, D., Kawanishi, Y., Ide, I., Murase, H.: Detector ensemble
  based on false positive mining for pedestrian detection. In: 2015 3rd IAPR
  Asian Conference on Pattern Recognition (ACPR). pp. 745--749. IEEE (2015)

\bibitem{Ref15}
Rehman, Y., Khan, J.A., Riaz, I., Shin, H.: Chunks: The remedy for notorious
  false alarms in pedestrian detection. In: 2016 International Conference on
  Electronics, Information, and Communications (ICEIC). pp. 1--4. IEEE (2016)

\bibitem{Ref16}
Krizhevsky, A., Sutskever, I., Hinton, G.E.: Imagenet classification with deep
  convolutional neural networks. Advances in neural information processing
  systems  25 (2012)

\bibitem{Ref17}
Ren, S., He, K., Girshick, R., Sun, J.: Faster r-cnn: Towards real-time object
  detection with region proposal networks. Advances in neural information
  processing systems  28 (2015)

\bibitem{Ref18}
Chen, L.C., Papandreou, G., Kokkinos, I., Murphy, K., Yuille, A.L.: Deeplab:
  Semantic image segmentation with deep convolutional nets, atrous convolution,
  and fully connected crfs. IEEE transactions on pattern analysis and machine
  intelligence  40(4),  834--848 (2017)

\bibitem{Ref19}
Szarvas, M., Yoshizawa, A., Yamamoto, M., Ogata, J.: Pedestrian detection with
  convolutional neural networks. In: IEEE Proceedings. Intelligent Vehicles
  Symposium, 2005. pp. 224--229. IEEE (2005)

\bibitem{Ref20}
Ouyang, W., Wang, X.: Joint deep learning for pedestrian detection. In:
  Proceedings of the IEEE international conference on computer vision. pp.
  2056--2063 (2013)

\bibitem{Ref21}
Zhang, L., Lin, L., Liang, X., He, K.: Is faster r-cnn doing well for
  pedestrian detection? In: Computer Vision--ECCV 2016: 14th European
  Conference, Amsterdam, The Netherlands, October 11-14, 2016, Proceedings,
  Part II 14. pp. 443--457. Springer (2016)

\bibitem{Ref22}
Cao, J., Pang, Y., Li, X.: Learning multilayer channel features for pedestrian
  detection. IEEE transactions on image processing  26(7),  3210--3220 (2017)

\bibitem{Ref23}
Konig, D., Adam, M., Jarvers, C., Layher, G., Neumann, H., Teutsch, M.: Fully
  convolutional region proposal networks for multispectral person detection.
  In: Proceedings of the IEEE conference on computer vision and pattern
  recognition workshops. pp. 49--56 (2017)

\bibitem{Ref24}
Brazil, G., Yin, X., Liu, X.: Illuminating pedestrians via simultaneous
  detection \& segmentation. In: Proceedings of the IEEE international
  conference on computer vision. pp. 4950--4959 (2017)

\bibitem{Ref25}
Kong, W., Li, N., Li, T.H., Li, G.: Deep pedestrian detection using contextual
  information and multi-level features. In: MultiMedia Modeling: 24th
  International Conference, MMM 2018, Bangkok, Thailand, February 5-7, 2018,
  Proceedings, Part I 24. pp. 166--177. Springer (2018)

\bibitem{Ref26}
Li, J., Liang, X., Shen, S., Xu, T., Feng, J., Yan, S.: Scale-aware fast r-cnn
  for pedestrian detection. IEEE transactions on Multimedia  20(4),  985--996
  (2017)

\bibitem{Ref27}
Zhou, C., Yuan, J.: Bi-box regression for pedestrian detection and occlusion
  estimation. In: Proceedings of the European Conference on Computer Vision
  (ECCV). pp. 135--151 (2018)

\bibitem{Ref28}
Zhang, S., Yang, J., Schiele, B.: Occluded pedestrian detection through guided
  attention in cnns. In: Proceedings of the IEEE conference on Computer Vision
  and Pattern Recognition. pp. 6995--7003 (2018)

\bibitem{Ref29}
Li, C., Song, D., Tong, R., Tang, M.: Illumination-aware faster r-cnn for
  robust multispectral pedestrian detection. Pattern Recognition  85,  161--171
  (2019)

\bibitem{Ref30}
Guan, D., Cao, Y., Yang, J., Cao, Y., Yang, M.Y.: Fusion of multispectral data
  through illumination-aware deep neural networks for pedestrian detection.
  Information Fusion  50,  148--157 (2019)

\bibitem{Ref31}
Gomez-Donoso, F., Cruz, E., Cazorla, M., Worrall, S., Nebot, E.: Using a 3d cnn
  for rejecting false positives on pedestrian detection. In: 2020 International
  Joint Conference on Neural Networks (IJCNN). pp. 1--6. IEEE (2020)

\bibitem{Ref32}
Iftikhar, S., Asim, M., Zhang, Z., El-Latif, A.A.A.: Advance generalization
  technique through 3d cnn to overcome the false positives pedestrian in
  autonomous vehicles. Telecommunication Systems  80(4),  545--557 (2022)

\bibitem{Ref33}
Intelligent multimodal pedestrian detection using hybrid metaheuristic
  optimization with deep learning model. Image and Vision Computing  131,
  104628 (2023)

\bibitem{Ref34}
Alfred~Daniel, J., Chandru~Vignesh, C., Muthu, B.A., Senthil~Kumar, R.,
  Sivaparthipan, C., Marin, C.E.M.: Fully convolutional neural networks for
  lidar--camera fusion for pedestrian detection in autonomous vehicle.
  Multimedia Tools and Applications  82(16),  25107--25130 (2023)

\bibitem{Ref35}
Liu, S., Huang, D., Wang, Y.: Adaptive nms: Refining pedestrian detection in a
  crowd. In: 2019 IEEE/CVF Conference on Computer Vision and Pattern
  Recognition (CVPR). pp. 6452--6461 (2019)

\bibitem{Ref36}
Huang, X., Ge, Z., Jie, Z., Yoshie, O.: Nms by representative region: Towards
  crowded pedestrian detection by proposal pairing. In: 2020 IEEE/CVF
  Conference on Computer Vision and Pattern Recognition (CVPR). pp.
  10747--10756 (2020)

\bibitem{Ref37}
Zhang, J., Lin, L., Zhu, J., Li, Y., Chen, Y.c., Hu, Y., Hoi, S.C.:
  Attribute-aware pedestrian detection in a crowd. IEEE Transactions on
  Multimedia  23,  3085--3097 (2020)

\bibitem{Ref38}
Tang, Y., Liu, M., Li, B., Wang, Y., Ouyang, W.: Otp-nms: Towards optimal
  threshold prediction of nms for crowded pedestrian detection. IEEE
  Transactions on Image Processing  (2023)

\bibitem{Ref40}
Jiao, Y., Yao, H., Xu, C.: Pen: Pose-embedding network for pedestrian
  detection. IEEE Transactions on Circuits and Systems for Video Technology
  31(3),  1150--1162 (2020)

\bibitem{Ref50}
Tan, M., Pang, R., Le, Q.V.: Efficientdet: Scalable and efficient object
  detection. In: Proceedings of the IEEE/CVF conference on computer vision and
  pattern recognition. pp. 10781--10790 (2020)

\bibitem{Ref43}
Dollar, P., Wojek, C., Schiele, B., Perona, P.: Pedestrian detection: An
  evaluation of the state of the art. IEEE transactions on pattern analysis and
  machine intelligence  34(4),  743--761 (2011)

\bibitem{Ref44}
Ouyang, W., Wang, X.: A discriminative deep model for pedestrian detection with
  occlusion handling. In: 2012 IEEE conference on computer vision and pattern
  recognition. pp. 3258--3265. IEEE (2012)

\bibitem{Ref45}
Zhang, S., Benenson, R., Schiele, B.: Citypersons: A diverse dataset for
  pedestrian detection. In: Proceedings of the IEEE conference on computer
  vision and pattern recognition. pp. 3213--3221 (2017)

\bibitem{Ref46}
Liu, W., Anguelov, D., Erhan, D., Szegedy, C., Reed, S., Fu, C.Y., Berg, A.C.:
  Ssd: Single shot multibox detector. In: Leibe, B., Matas, J., Sebe, N.,
  Welling, M. (eds.) Computer Vision -- ECCV 2016. pp. 21--37. Springer
  International Publishing, Cham (2016)

\bibitem{Ref47}
Redmon, J., Farhadi, A.: Yolov3: An incremental improvement. CoRR
  abs/1804.02767 (2018), \url{http://arxiv.org/abs/1804.02767}

\bibitem{Ref48}
Lin, T.Y., Dollár, P., Girshick, R., He, K., Hariharan, B., Belongie, S.:
  Feature pyramid networks for object detection. In: 2017 IEEE Conference on
  Computer Vision and Pattern Recognition (CVPR). pp. 936--944 (2017)

\bibitem{Ref49}
Wang, R.J., Li, X., Ling, C.X.: Pelee: a real-time object detection system on
  mobile devices. In: Proceedings of the 32nd International Conference on
  Neural Information Processing Systems. p. 1967–1976. NIPS'18, Curran
  Associates Inc., Red Hook, NY, USA (2018)

\bibitem{Ref51}
Liu, Z., Lin, Y., Cao, Y., Hu, H., Wei, Y., Zhang, Z., Lin, S., Guo, B.: Swin
  transformer: Hierarchical vision transformer using shifted windows. In:
  Proceedings of the IEEE/CVF international conference on computer vision. pp.
  10012--10022 (2021)

\end{thebibliography}

\end{document}